\documentclass{amia}
\usepackage{graphicx}
\usepackage[labelfont=bf]{caption}
\usepackage[superscript,nomove]{cite}
\usepackage{color}
\usepackage{caption}
\usepackage{subcaption}
\usepackage{booktabs}
\usepackage{siunitx}
\usepackage[english]{babel}
\usepackage[utf8x]{inputenc}
\usepackage[T1]{fontenc}
\usepackage{amsmath}
\usepackage{listings}
\usepackage{amsmath}
\usepackage{graphicx}
\usepackage[colorinlistoftodos]{todonotes}
\usepackage{amsthm}
\usepackage{amsfonts}
\usepackage{amssymb}
\usepackage{bm}
\usepackage{cleveref}
\usepackage{booktabs}
\usepackage{multirow}
\usepackage{multicol}
\usepackage{comment}
\usepackage{graphicx}
\usepackage{subcaption}
\usepackage{amssymb}
\usepackage{algpseudocode}
\usepackage{algorithm}
\usepackage{outlines}
\usepackage{textcomp}
\usepackage{moreverb,url}
\usepackage{pbox}
\usepackage{color, soul}
\usepackage[flushleft]{threeparttable}
\usepackage{xcolor}

\begin{document}

\title{Identifying Opioid Use Disorder from Longitudinal Healthcare Data using a Multi-stream Transformer}

\author{Sajjad Fouladvand, MSc$^{1,2}$, Jeffery Talbert, PhD$^{1,3}$, Linda P. Dwoskin, PhD$^{4}$, \\Heather Bush, PhD$^{5}$, Amy Lynn Meadows, MD $^{6}$, Lars E. Peterson, MD, PhD $^{7,8}$,  Steve K. Roggenkamp, MSc$^{1}$, Ramakanth Kavuluru, PhD$^{1,2,3}$, Jin Chen, PhD$^{1,2,3}$}

\institutes{
    $^1$Institute for Biomedical Informatics; $^2$Department of Computer Science;
    $^3$Department of Internal Medicine;
 $^4$Department of Pharmaceutical Sciences;
 $^5$Department of Biostatistics;
 $^6$Department of Psychiatry;
 $^7$Department of Family and Community Medicine, University of Kentucky, Lexington, KY, USA
 $^8$American Board of Family Medicine, Lexington, KY, USA;
  \\
}

\maketitle

Note: This manuscript has been accepted by AMIA 2021 for oral presentation on November 1, 2021.

\noindent{\bf Abstract}

\textit{Opioid Use Disorder (OUD) is a public health crisis costing the US billions of dollars annually in healthcare, lost workplace productivity, and crime. Analyzing longitudinal healthcare data is critical in addressing many real-world problems in healthcare. Leveraging the real-world longitudinal healthcare data, we propose a novel multi-stream transformer model called MUPOD for OUD identification. MUPOD is designed to simultaneously analyze multiple types of healthcare data streams, such as medications and diagnoses, by attending to segments within and across these data streams. Our model tested on the data from 392,492 patients with long-term back pain problems showed significantly better performance than the traditional models and recently developed deep learning models. } 

\section{Introduction}
Early identification and engagement of individuals at risk of developing an opioid use disorder (OUD) is a critical unmet need in healthcare~\cite{gostin2017reframing, pitt2018modeling}. Individuals with OUD often do not seek treatment or have internalized stigma about OUD that limits identification through traditional means, such as screening and clinical interview~\cite{olsen2014confronting}. Significant disparities limit access to treatment for OUD resulting in less than 20\% of all individuals with OUD receiving any form of treatment in the past year~\cite{wu2016treatment}. 
While there are currently tools developed to predict aberrant behavior when prescribing opioids\cite{oudriskaber} or to predict OUD from a general primary care population\cite{gao2021predicting}, there are only a few clinical tools, such as the Opioid Risk Tool\cite{webster2005predicting}, developed for assessing the risk of OUD. Typical clinician workflow does not allow for comprehensive OUD screening, but available administrative and clinical data have the potential to help clinicians identify and screen higher risk patients providing an opportunity for primary care professionals to play a greater role in increasing OUD detection, treatment, and prevention.
Healthcare data are a growing source of information that can be harnessed together with machine learning to advance our understanding of factors that increase the propensity for developing OUDs as well as those that aid in the treatment of the disorders\cite{RN4, segal2020development}. In healthcare data, patients' outcomes and  treatments are collected at multiple follow-up times.
%
Tools developed to analyze longitudinal healthcare data and to extract meaningful patterns from these ever growing data are critical in addressing real-world public health emergency including but not limited to OUD. 

Analyzing real-world data is a complicated task with multiple computational challenges including high dimensionality, heterogeneity, temporal dependency, sparsity, and irregularity~\cite{RN13}. In particular, healthcare (and claim) data are typically collected from multiple sources, and the subsequent data analysis requires simultaneous analysis of the temporal correlation among multiple streams such as medications, diagnoses, and procedures. 
Deep learning models have demonstrated great potential in addressing some of these challenges and creating promising longitudinal healthcare data analysis tools. 
Among them, Doctor AI~\cite{choi2016doctor}, RETAIN~\cite{choi2016retain}, and DeepCare~\cite{pham2016deepcare} modeled multiple data streams including medications, diagnoses, and procedures using Recurrent Neural Network (RNN) models such as Long-Short Term Memory models (LSTMs)~\cite{RN17}. 
Doctor AI concatenated multi-hot input vectors to predict subsequent visit events~\cite{choi2016doctor}. RETAIN used two separated RNNs to generate attentions at the visit level and the variable level as well~\cite{choi2016retain}. 
%
These applications demonstrate that RNNs are promising in longitudinal and sequential healthcare data analysis, 
since RNNs are capable of extracting contextual information from past time steps and pass this information forward; this helps to efficiently model long-term dependencies in input streams\cite{RN25}. 
Nevertheless, the network architecture and design preclude RNNs from processing long streams in a reasonable amount of time~\cite{ma2017dipole}. Attention mechanism was introduced in RNNs to increase their capacity in capturing long range dependencies more efficiently\cite{ma2017dipole, luong2015, bahdanau2014}. Attention-based models bridge the gap between different states in RNNs using a context vector. Successful applications of multiple attention
layers led to the transformer model~\cite{NIPS2017_7181}, which removed recurrence in RNNs relying entirely on the attention mechanism.  

The transformer is a type of attention-based deep learning models originally proposed for natural language processing (NLP) tasks such as machine translation~\cite{NIPS2017_7181}. Later, transformers have been applied on longitudinal EHR data~\cite{choi2020learning} to predict patients' outcomes in the future. There are already several models that have been successfully applied on EHR data without significantly changing the network architecture or loss~\cite{song2018attend, wang2019inpatient2vec, shang2019pre}. 
Of course, the typical transformer\textquotesingle s structure can be altered to better fit the special needs of solving healthcare problems~\cite{choi2020learning,li2020behrt}.  
Choi et al. proposed a transformer model for healthcare data analysis by utilizing the conditional probabilities calculated from the encounter records to guide the self-attention mechanism in the transformer~\cite{choi2020learning}. 
BEHRT~\cite{li2020behrt} was developed based on BERT~\cite{bert}, a popular transformer model for NLP tasks, for analyzing EHR data. BEHRT considers the patients' existing diagnoses and demographic data to predict their future diagnoses. 
%
%
Similar to RNNs, transformers have been modified to model multiple data streams. Li et al developed a two-stream transformer to analyze both time-over-channel
and channel-over-time features in human activity recognition tasks~\cite{bing2021that}. Two parallel, yet separate transformers were used to handle two input streams. 
Another multi-stream transformer has been developed to generate effective self-attentions for speech recognition~\cite{han2019state}. They parallelized multiple self-attention encoders to process different input speech frames. 
Gomez et al. developed a multi-channel transformer for sign language translation using one self-attention encoder~\cite{camgoz2020multi}. Their model finds the attentions across three different channels, i.e. hand shapes, mouthing, and upper body pose. 
A more recent work~\cite{hu2021transformer} showed that ``transformer is all you need'' by using multiple transformer encoders. The encoded outputs can be concatenated using a joint decoder that enables simultaneous model training. 
There are also works that analyze multi-stream data using transformer by simply stacking or parallelizing multiple transformer models~\cite{libovicky2018input, li2019visualbert}.

\begin{figure}[!bt]
	\centering
	\includegraphics[width=0.9\textwidth]{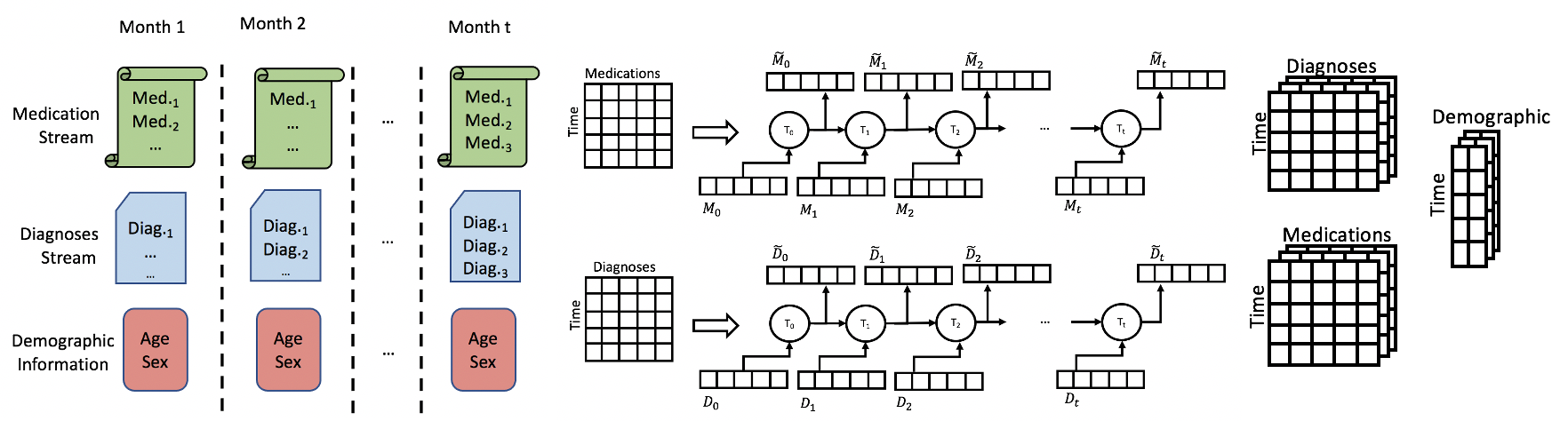}
	\caption{Data preprocessing and patient representation. EHR data are first converted to an enrollee-time matrix $X(P,T,F)$. Then, the data are fed to LSTM models to encode the medication and diagnosis streams separately. }
	\label{fig:data_format}
\end{figure}

Although the recently developed transformer models showed promising performance, especially on handling multiple data streams, the potential of applying transformers on healthcare data analysis has not yet been fully explored. One of the major limitations is the lack of capacity to model multiple data streams within the self-attention layer. 
The transformer was originally designed to process one data stream, which is mostly an order of words in a NLP task, at a time. The modified transformers either can only handle multiple streams at intra-stream level or they are not suitable to solve  OUD identification problem as a real-time task where only previous clinical events can be used to make a decision at a specific time point.
Here, OUD identification is a complex data analysis task that includes not only finding long term effects of prescription opioids such as morphine and fentanyl, history of diagnoses such as mood disorders, but also the hidden associations between patient's prescriptions and diagnoses, since these input streams are highly correlated with each other. Identifying the relationships within and between input streams may reveal hidden patterns leading to an increased classification ability and interpretabilty for OUDs. Moreover, the medication application patterns and the interactions between medications across different visits as well as the patient's diagnoses patterns throughout his/her medical history may carry important information that should be extracted in order to develop precise and sensitive OUD identification tools. 

This study proposes a novel transformer model called {\bf Mu}lti-stream Transformer for {\bf P}redicting {\bf O}pioid Use {\bf D}isorder (MUPOD) to analyze longitudinal healthcare data collected from multiple sources and predict the onset of OUD. First of all, MUPOD is capable of analyzing multiple data streams, such as medication and diagnosis, simultaneously and extracting associations within and between the streams. Second, MUPOD utilizes attention weights within and across data streams to interpret the classification results. 
In our experiment, MUPOD successfully captured the complex associations within and between multiple streams including medications, diagnoses, and demographic information, and predicts the onset of OUD precisely. 

\section{Materials and Methods}
\subsection*{Data Set}\label{dataset}

The large-scale administrative records in the IBM (formerly Truven Health Analytics) MarketScan Commercial Claims~\cite{truven} database were used to train and test both baseline models and MUPOD. Data include person-specific clinical utilization, expenditures and enrollment across inpatient, outpatient, prescription drug and carve-out services. The database contains about 30 million enrollees, a nationally representative sample of the US population with respect to sex (50\% female), regional distribution, and age.

We extracted medications, diagnoses and demographic information of 682,402 patients who have at least one diagnosis of OUD (ICD-9: 304.0x, 305.5x and ICD-10: F11.xxx; where x can be any code) from 2009 to 2018. 
%
The hyper-geometric~\cite{rice2006mathematical} test was used to identify sub-cohorts of OUD with high statistical significance of whether a population consists the richest information of OUD. We identified an OUD sub-cohort (p-value 0.00) with 229,214 patients who had at least one Clinical Classification Software code (CCS) of 205 (patients with Spondylosis; intervertebral disc disorders; other back problems). This sub-cohort was defined as  the case cohort. Note that CCS 205 has already been shown to be a prevalent diagnosis in OUD patients in the literature~\cite{mark2013psychiatric, barnett2009comparison}.  

The case cohort (OUD positive and CCS 205) was matched with a subpopulation of OUD-negative patients called the control cohort. All the individuals in the control cohort have the same back pain diagnosis (CCS 205) but do not develop OUD. 
We first matched cases and controls based on age and sex. 
Second, we matched them based on the opioid medication use duration. Specifically, we grouped every opioid medication with a therapeutic class generic
product identifier (TCGPI) of 65x as opioid medications. Buprenorphine and Methadone were excluded as they are often used as a treatment for opioid overdose. Next, we randomly sampled OUD-negative patients who have the matched age and gender with the case ensuring that the averaged opioid use ratio between case and control is almost equal. 

Table~\ref{tab:data_basic_stat} shows the characteristics of cases and controls regarding age, sex, top-10 most frequent medications and top-10 most frequent diagnoses. 
The diagnoses and the medications were classified using CCS codes and Generic Product Identifier codes (TCGPI) respectively. We grouped opioid analgesics, anticonvulsants (neuromuscular agents), musculoskeletal therapy agents,
and antianxiety agents based on the first two digits of their TCGPI codes as 65x, 72x, 75x, and 57x, respectively. The rest of medications were classified using the first 6 digits of their TCGPI codes (from left to right). The variables presented in Table~\ref{tab:data_basic_stat} have already been reported as OUD risk factors in the literature~\cite{mark2013psychiatric, clarke2014rates}. Especially, diseases including ``Other connective tissue disease'', ``Other nervous system disorders'', ``Essential hypertension'', ``Mood disorders'', ``Other non-traumatic joint disorders and Anxiety disorders'' have been found to be more prevalent diagnoses among OUD patients than normal people~\cite{mark2013psychiatric}. Note, since we matched the case and control cohorts based on age, sex and analgesics-opioid use, these three variables have similar statistical characteristics across both case and control cohorts. However, the distributions of other variables vary across the case and control cohorts and can be utilized by our deep learning models to discriminate OUD-positive patients from OUD-negative individuals.
 

\begin{table}[!bt]
	\centering 
	\caption{Distributions of age, sex, medication, and diagnoses in case and control patients. Top 10 diagnoses and medications are provided. The numbers indicate the number of patients who had at least one such diagnosis or medication.}
	\begin{tabular}{p{120pt}p{40pt}p{40pt}|p{120pt}p{40pt}p{40pt}}
		\toprule
		{\textbf{Variables}}& \textbf{Case} & \textbf{Control} & {\textbf{Variables}}& \textbf{Case} & \textbf{Control}\\
		\toprule
		{\textbf{Demographics}}&  & \\    
		\toprule
		{Age (SD)} & 45.62 (13.81) & 52.35 (14.39)  &		
		{Female (percentage)}& 109,121 (55.60\%) & 117,699 (59.98\%) \\
		\midrule		
		{\textbf{Diagnoses (CCS Code)}}&  & & {\textbf{Medications (TCGPI Code)}}& & \\
		\toprule
		{Other connective tissue disease (211)}& 152,703 (77.81\%)  & 165,112 (84.14\%)&Analgesics - Opioid (65) &190,141 (96.89\%)&196,246 (100\%) \\
		\midrule
		{Other nervous system disorders (95)}&138,866 (70.76\%) &141,350 (72.03\%) &Neuromuscular Agents Anticonvulsants (72)&105,508 (53.76\%)& 97,444 (49.65\%)\\
		\midrule
		{Essential hypertension (98)}& 106,299 (54.17\%)& 132,049 (67.29) &Musculoskeletal Therapy Agents (75)&106,186 (54.11\%)&102,888 (52.43\%)\\
		\midrule
		{Mood disorders (657)}& 97,035 (49.45\%)& 81,306 (41.43\%) &Antianxiety Agents  (57)&76,830 (39.15\%)&  75,463 (38.45\%)\\
		\midrule
		{Other aftercare (257)} &  127,131 (64.78\%)& 133,920 (68.24\%) &{Proton Pump Inhibitors (492700)}& 71,243 (36.30\%)& 86,561 (44.11\%)\\
		\midrule
		{Residual codes; unclassified (259)}& 136,177 (69.39\%)&152,748 (77.83\%) &{Serotonin-norepinephrine Reuptake Inhibitors (581800)}&58,039 (29.57\%)& 48,323 (24.62\%)\\
		\midrule
		{Other non-traumatic joint disorders (204)}&  134,042 (68.30\%)& 150,660 (76.77\%)&{Selective Serotonin Reuptake Inhibitors (581600)}&69,665 (35.50\%)& 65,005 (33.12\%)\\
		\midrule
		{Anxiety disorders (651)}&  91,736 (46.75\%)& 78,296 (39.90\%) &{Hmg Coa Reductase Inhibitors (394000)}&53,806 (27.42\%)&  79,201 (40.36)\\
		\midrule
		{Disorders of lipid metabolism (53)}&94,507 (48.16\%)&122,322 (62.33\%) &{Non-barbiturate Hypnotics (602040)}&46,965 (23.93\%)& 44,404 (22.63\%)\\
		\midrule
		{Medical examination/evaluation (256)}&129,224 (65.85\%) &147,268 (75.04\%)&{Nonsteroidal Anti-inflammatory Agents (661000)}&87,301 (44.49\%)& 98,639 (50.26\%)\\
		\bottomrule
	\end{tabular}%
	\label{tab:data_basic_stat}%
\end{table}%

\subsection*{Data Pre-processing}

For each of the enrollees in the case and control cohort, his/her medications and diagnoses between Jan 2009 and Dec 2018 and demographic records were extracted. In total, we extracted 78,136,935 medication records and 143,275,864 diagnoses records. The original format of the prescription and professional service encounter claims in IBM MarketScan data is a table where each row is a visit and columns are enrollee ID, date of visit, and prescription/diagnoses. If an enrollee has multiple visits, each visit will occupy a row in the table. To facilitate further study of the temporal patterns in the data, we converted the data into an enrollee-time matrix $X (P, T, F) $ where each $x_{i,j} \subseteq F$ is a set of medications or  diagnoses (from feature space $F$) associated with enrollee $p_i \in P$ at time slot $t_j \in T$, where $P$ is the enrollee set and $T$ is the set of monthly slots between Jan 2009 and Dec 2018. We excluded patients from $X(P, T, F)$ if the number of valid entries is less than 3. 
%



The goal of data representation is to learn a function: $f_R: X \rightarrow \mathbb{R}^d$, 
%
where $d$ is 10 in this work and it shows the dimension of the representation to which each input stream is mapped, $X \in \{M, D\}$, and $M$ and $D$ are medication and diagnosis, respectively. To train the function $f_R$, LSTM~\cite{sud2019Sajjad,RN19} was adopted. The outputs from all LSTM hidden states were used to represent both the OUD case and control cohorts. The general schema of the data pre-processing and representation is shown in Figure~\ref{fig:data_format}. 	 

\subsection*{MUPOD Architecture}

MUPOD is a transformer-based deep learning model designed to analyze $n$ highly correlated healthcare data streams simultaneously. 
To minimize ambiguity, the algorithm is described for a single patient and for $n=3$. Each patient can be represented by $p=(S, y)$ in which $S$ is a set of input streams and $y$ is the target label. Herein, three input streams are considered: 1) medication tuples $(T, M)$ in which $t_i$ is the $i^{th}$ time step and $M$ is a list of medications that the patient is prescribed with at time $t_i$, 2), diagnoses tuples $(T, D)$ where $t_i$ is the $i^{th}$ time step and $D$ is a list of diagnoses assigned to the patient $P$ at time $t_i$, 3) demographic tuples $(T, G)$ in which $t_i$ is the $i^{th}$ time step and $G$ is the demographic information of patient $P$ at $t_i$. 

This study uses the encoder part of transformer to identify the associations between medication and diagnosis across time and detect the onset of OUD. 
Medications $M$, diagnoses $D$, and demographics $G$ are fed to the model in parallel. The first step is to incorporate the temporal patterns of the data stream into the encoder's inputs using positional encoding. The embedding layer in the transformer is replaced by the proposed LSTM based representation layer. This change has two computational advantages. Firstly, it deals with challenges in the input data such as variable dimension and data sparseness, which is common in longitudinal healthcare data. Secondly, it extracts hidden parameters and transforms the original input into a new feature space where cases and controls are better separated than in the original feature space.

The encoded input streams are plugged into the attention layer to generate Query, Key, and Value matrices for each input stream. For example, medications $M$ are fed to a set of fully connected layers to generate $M_Q$, $M_K$, and $M_V$, representing the query, key, and value matrices for the encoded medication stream for patient $P$. Let $X, Y \in \{M, D\}$, the Query, Key, and Value matrices are used to find the attentions across these three input streams:
\begin{equation} \label{eq:my_att}
\begin{split}
    Attention (X_Q, Y_K, Y_V)= & \;softmax(\frac{X_Q Y_K^\top}{\sqrt{d_k}}) Y_V  
\end{split}
\end{equation}

\begin{figure}[!bt]
	\centering
	\begin{subfigure}{.4\linewidth}
		\centering
		\includegraphics[width=.7\linewidth]{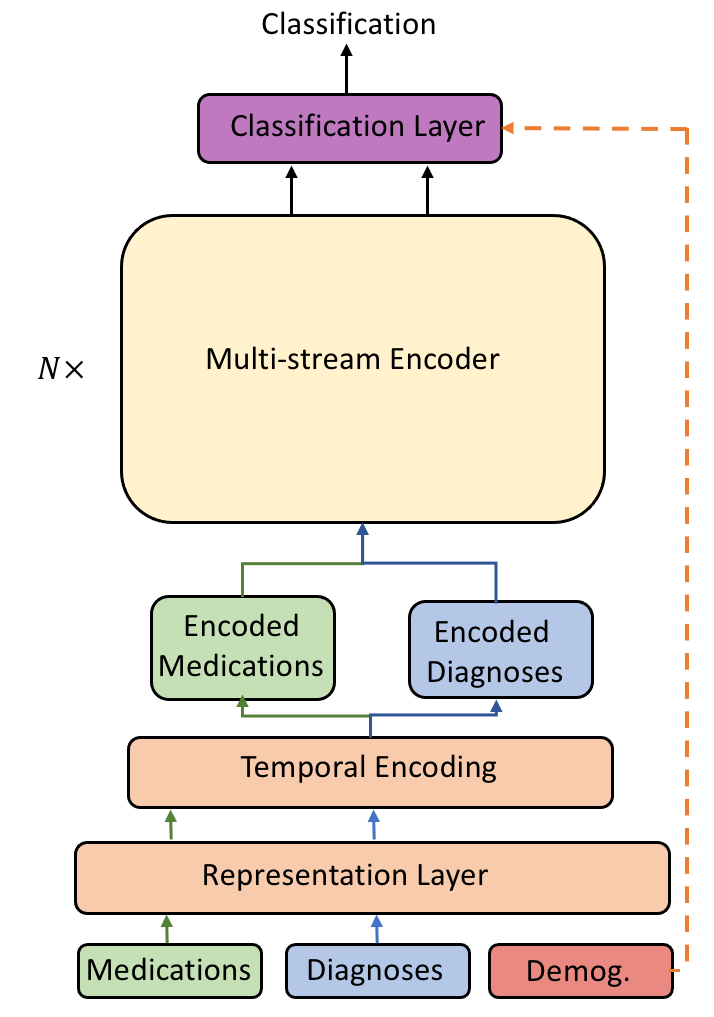}
		\caption{General architecture of MUPOD.}
		\label{fig:model_arch_a}
	\end{subfigure}%
	\begin{subfigure}{.5\linewidth}
		\centering
		\includegraphics[width=.7\linewidth]{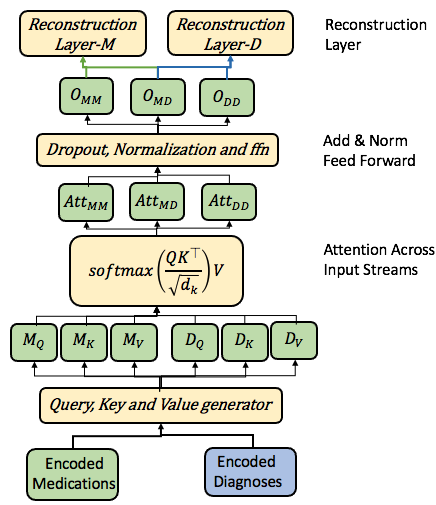}
		\caption{Multi-stream encoder.}
		\label{fig:model_arch_b}
	\end{subfigure}
	\caption{MUPOD architecture. $X_Q, X_K, and \quad X_V $ represent query, key, and value matrices for the input stream $X$, where $X \in \{Medication, Diagnoses\}$. $Att_{XY}$ represents the attention weights between different records across input streams $X$ and $Y$, where $X, Y \in \{Medications, Diagnoses\}$. $O_{XY}$ represents the outputs, which capture the associations between the input streams $X$ and $Y$. The demographic information is plugged into the system before the last layer and in the classification layer.}
	\label{fig:model_arch}
\end{figure}

Note, the $d_k$ is the same as the original transformer. Figure~\ref{fig:model_arch_a} describes how the data flows through the different layers of MUPOD. The raw medication and diagnose streams are first represented in the representation layer (the intermediate outputs of the LSTMS models in Figure~\ref{fig:data_format}). The temporal information is then encoded into the represented streams in the temporal encoding layer. The encoded streams are processed in the MUPOD's multi-stream encoder layer. This novel multi-attention layer is further described in more details in Figure~\ref{fig:model_arch_b}. In the figure, $X_Q$, $X_K$, and $X_V$ represent query, key, and value matrices for stream $X$ ($X \in \{M, D\}$). All possible combinations of the streams are used to determine the attention weights between different visits and across streams. Attentions are then passed through a set of dense layers to generate outputs. For example, given two data streams $M$ and $D$, we can generate three combinations i.e. $MM$, $MD$, and $DD$. 

The reconstruction layer receives the relevant outputs and maps them to appropriate format for the next layer as described in Equation~\ref{eq:recon_layer}. For example, only the outputs relevant to the medications ($M$) including $O_{MM}$ and $O_{MD}$ are used to reconstruct the medication stream appropriate to be fed into the next encoder layer:
\begin{equation} \label{eq:recon_layer}
\begin{split}
	f: & \; O_{XX}, O_{XY}  \longrightarrow \hat{X}\\
    \hat{X}= & [Concat (O_{XX}, O_{XY})]W_x + b_x 
\end{split}
\end{equation}
where $ O_{XX}, O_{XY} \subset \{O_{MM}, O_{MD}, O_{DD}\}$, $  \hat{X} \in \{\hat{M}, \hat{D}\}$, $X, Y \in \{M, D\}$, $W_x$ and $b_x$ are trainable reconstruction weight and bias matrices. The two reconstructed matrices generated by the last encoder layer are fed to classification
layer to make the final decision for the current patient p as $Softmax([Concat(\hat{M}, \hat{D})] W + b)$.

\section*{Experimental Results}

All the deep learning models in this work were deployed on the TensorFlow platform~\cite{RN21} and were trained using eight GeForce GTX 1080 GPUs. The original transformer model, LSTM models, Linear Regression (LR), Random Forest (RF)~\cite{RN15} and Support Vector Machine (SVM)\cite{RN14} were compared with MUPOD as baselines. We used 314,504 samples for training, 38,776 samples for validation and 39,212 for testing the models. All results reported in this paper are on the test set. We optimized all models using a random search policy across hyper-parameters of each model. A grid of hyper-parameters values was set up and 10 random combinations of the hyper-parameters were selected to train the models. 

The optimized SVM model uses a RBF kernel function and the optimum value for the parameter $C$ is 0.0039 in this model. The optimized linear regression model uses the L2 norm with $C=0.0625$ in penalization and the sag algorithm as it\textquotesingle s solver method. The optimum number of trees in the random forest model is 1600 and the optimum value for maximum number of levels in a tree is 40. For the LSTMs, their learning rates were randomly set to $10^n$ where $ n \in \{-2, -3, -4\} $. The batch size was randomly selected from $\{64, 256, 512\}$ and the number of iterations was randomly selected from $n \times 10^3$ where $n \in \{10, 50, 100, 200\}$. The regularization parameter for LSTM models was randomly selected from $10^{n}$ where $n \in \{-4, -5, -6\}$. The number of hidden neurons for the LSTMs in the representation layer was fixed to 10; because the outputs of these LSTM models were the inputs to MUPOD and the inputs to our model have to be of a fixed dimension (the dimension of our model in this paper is 20: 10 for medications and 10 for diagnoses stream). However, the number of hidden neurons for the other LSTM model used as a baseline (refer to Table~\ref{tab:oud_prediction_dataset_S2}) was randomly selected from $2^n$ where $n \in \{3, 4, 5, 6, 7, 8\}$.

Table~\ref{tab:oud_prediction_dataset_S2} compares the classification performance of MUPOD with LR, RF, SVM, LSTM and the original Transformer model. We used the same train, validation and test data to train, validate and test all models in Table~\ref{tab:oud_prediction_dataset_S2} except for the SVM model. Due to the hardware and time limitation we had to train and test this model using 10,000 randomly selected samples. Note, the LSTM model in Table \ref{tab:oud_prediction_dataset_S2} is trained using medication, diagnosis and demographic data. We concatenated the vectors of medication, diagnosis and demographics in each time step and formed a single vector which was fed to this LSTM model. We dynamically unrolled the LSTM model based on the input sequences\textquotesingle \,lengths and applied a fully connected layer and an argmax function on the last output of the unrolled LSTM model to make the final decisions. The hyper-parameter search space for this LSTM was the same as explained earlier in this section. We used a randomized 5-fold cross validation to tune LR, RF and SVM models. The LR, RF and SVM were trained on the static data and the LSTM, transformer and MUPOD were trained on the longitudinal data. To create static data for LR, RF and SVM, the longitudinal data was converted to a new format $Y(P, L)$, where $P$ is the complete list of patients, and $L$ is a vector including aggregated values for all medication, diagnosis and demographic features across time steps (from Jan. 2009 to Dec. 2018). In fact, we counted the frequencies for each medications and diagnoses and concatenated these frequencies with demographic information  of the patients  to create $L$. Transformer is the original encoder block of the transformer model~\cite{NIPS2017_7181}. We concatenated the vectors of medication, diagnosis and demographics and fed them to the original encoder block of the transformer model. Then, a fully connected layer and softmax function were used to perform the final classifications. In Table \ref{tab:oud_prediction_dataset_S2}, MUPOD has the highest accuracy (0.775), precision (0.741), F1-score (0.790) and AUC (0.871). These results indicate that our proposed model captures important factors in the medication, diagnosis and demographic data and provides an increased power to detect the development of OUD, while LR, RF, SVM, LSTM and original Transformer appear to miss such factors. 

\begin{table}[!bt]
	\centering \vspace{0.1in}
	\caption{Performance of OUD classification using MUPOD compared to RF, SVM, LSTM and original transformer.} \vspace{-0.1in}
	\begin{tabular}{p{80pt}p{45pt}p{45pt}p{45pt}p{45pt}p{45pt}p{60pt}}
		\toprule
		{\textbf{Model}} & \textbf{Acc.}&
		{\textbf{Prec.}} & {\textbf{Rec.}} & {\textbf{F1-score}} & {\textbf{AUC}}& {\textbf{P@R=.8$\pm$0.001}}\\
		\midrule
		{LR} & 0.638 & 0.641  & 0.625 & 0.633 & 0.689 & 0.463 \\	
		\midrule
		{RF} & 0.698 & 0.693  & 0.710 & 0.702 & 0.774 & 0.449\\	
		\midrule
		{SVM} & 0.569 & 0.539  & 0.831 & 0.654 & 0.677 & 0.478\\	
		\midrule	
		{LSTM} & 0.693 & 0.784  & 0.533 & 0.635 & 0.790 & 0.666\\	
		\midrule	
		{Transformer} & 0.708 & 0.654  & \textbf{0.880} & 0.751 & 0.801 & 0.689\\	
		\midrule		
		{MUPOD} & \textbf{0.775} & \textbf{0.741} & 0.847  & \textbf{0.790}  &\textbf{ 0.871} &0.771\\
		\bottomrule
	\end{tabular}%
	\label{tab:oud_prediction_dataset_S2}%
\end{table}

In addition, we tested the models\textquotesingle \,performances on three imbalanced test data sets with the ratio of OUD-positive samples to OUD-negative samples set to 0.1, 0.2 and 0.5. OUD is an uncommon event and the ratio of OUD-positive to OUD-negative patients in patients who have used Opioid prescriptions at least 3 times is 3.2\% in the data set. Therefore, we conducted the experiments in Table~\ref{tab:oud_prediction_imb} to simulate the performance of the models on imbalanced datasets as well. Table~\ref{tab:oud_prediction_imb} shows the model performances on imbalanced test sets. Table~\ref{tab:oud_prediction_imb}  shows  that  MUPOD maintains  higher performance on all imbalanced test sets compared to all baselines in terms of precision, F1-score and AUC.  Note, we did not show accuracy in Table~\ref{tab:oud_prediction_imb}, because this measure is not informative when assessing algorithms on imbalanced data.

\begin{table}[!bt]
	\centering \vspace{0.1in}
	\caption{OUD classification results for imbalanced test sets. The $.xN$ means the number of samples in the OUD-positive cohort are $0.x$ times smaller than the number of samples in the OUD-negative cohort.} \vspace{-0.1in}
	\begin{tabular}{p{40pt}p{1pt}p{12pt}p{12pt}p{12pt}p{.01pt}p{12pt}p{12pt}p{12pt}p{.01pt}p{12pt}p{12pt}p{12pt}p{.01pt}p{12pt}p{12pt}p{12pt}p{.01pt}p{12pt}p{12pt}p{12pt}}
		\toprule
		\multirow{2}{*}{Model} &
		\multicolumn{5}{c}{Precision} &
		\multicolumn{4}{c}{Recall} &
		\multicolumn{4}{c}{F1-score} &
		\multicolumn{3}{c}{AUC} \\ \cline{3-5} \cline{7-9} \cline{11-13} \cline{15-17} 
		& {}& {$.5N$} & {$.2N$} & {$.1N$} &{}& {$.5N$} & {$.2N$} & {$.1N$}&{}& {$.5N$} & {$.2N$} & {$.1N$}&{}& {$.5N$} & {$.2N$} & {$.1N$}\\ 
		\midrule
		RF & &  .531 & .313& .182 & & .710 & .715 & .701& & .608 & .436& .290 &  & .773 & .777 & .770  \\
		LSTM & & .539 & .312& .189 & & .548 & .532 & .546& & .544 & .393 & .281 &  & .730 & .723 & .732  \\		
		Transformer & & .486 & .276& .160  && \textbf{.879} & \textbf{.885}& \textbf{.883} & & .626 & .420& .270  && .799 & .804 & .796\\
		MUPOD & & \textbf{.588} & \textbf{.364} & \textbf{.221} && .845 & .848& .843  && \textbf{.693} & \textbf{.509}& \textbf{.351}  & &\textbf{.871} & \textbf{.870} & \textbf{.871}\\
		\bottomrule
	\end{tabular}
\label{tab:oud_prediction_imb}%
\end{table}



We examined the relationships between the medication and diagnosis streams by aggregating the attention weights in the first layer of the model for all the records of each individual and visualized the results. While it is still unclear whether attentions can be used to explain deep learning models~\cite{serrano2019attention, jain2019attention}, attention weights have been used extensively to assess feature importance~\cite{wiegreffe2019attention, li2020behrt}. In particular, the aggregated attentions across all the records of the same patient may be useful to identify important relationships between his/her prescriptions and diagnoses. In the visualization, a rectangular node represents a medication type and an oval node represents a diagnosis code. We divided the accumulated attention weights to ``moderate'' and ``strong'' based on pre-defined thresholds (i.e. moderate: $0.3\sim0.6$, and strong: $\geqslant0.6$) that were selected by visually inspecting the distribution of accumulated attention weights. The moderate and strong connections are represented using dashed and solid lines respectively. The lines of an OUD-negative patient are colored black, while the lines of an OUD-positive patient are colored red. 

\begin{figure}[!bt]
	\centering
	\begin{subfigure}{.33\linewidth}
	\centering
	\includegraphics[width=.7\linewidth]{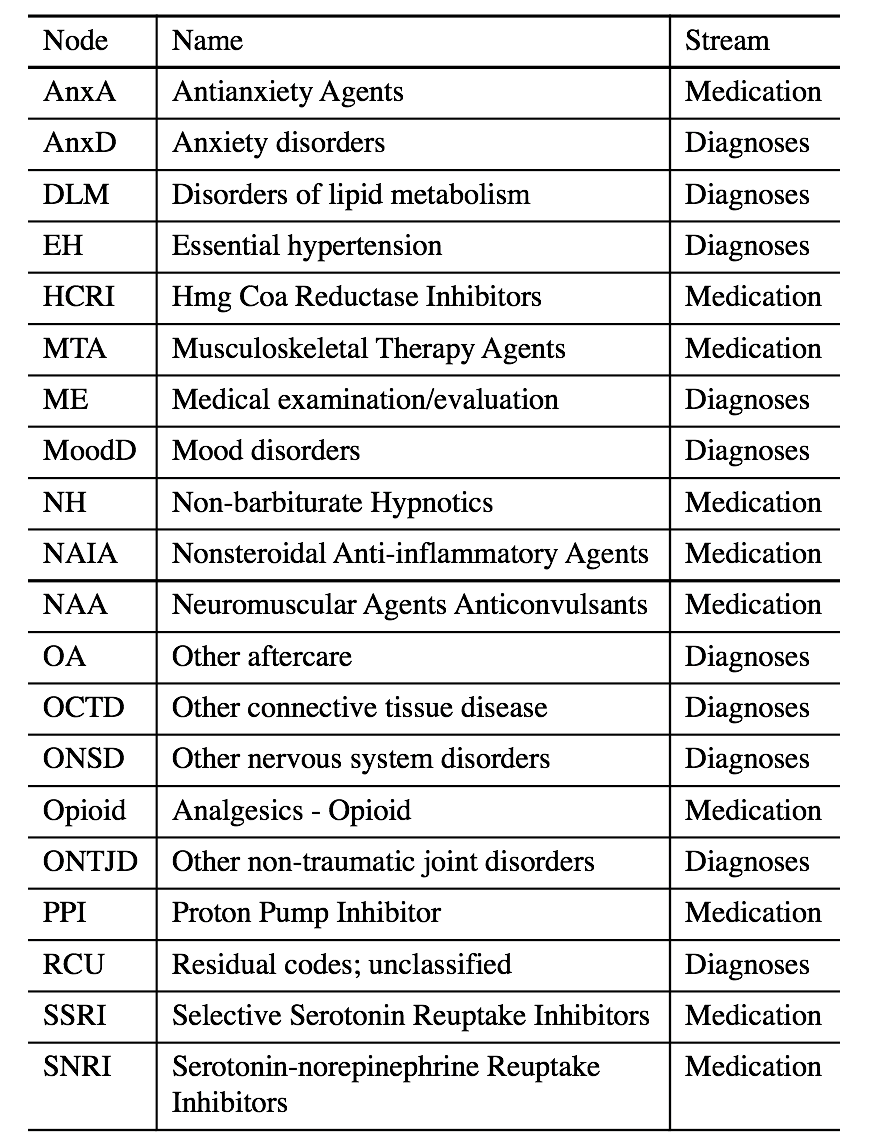}
	\caption{Medication and diagnoses abbreviations and full names.}
	\label{fig:attentions_md_legend}
\end{subfigure}\hfill%
	\begin{subfigure}{.33\linewidth}
		\centering
		\includegraphics[width=.7\linewidth]{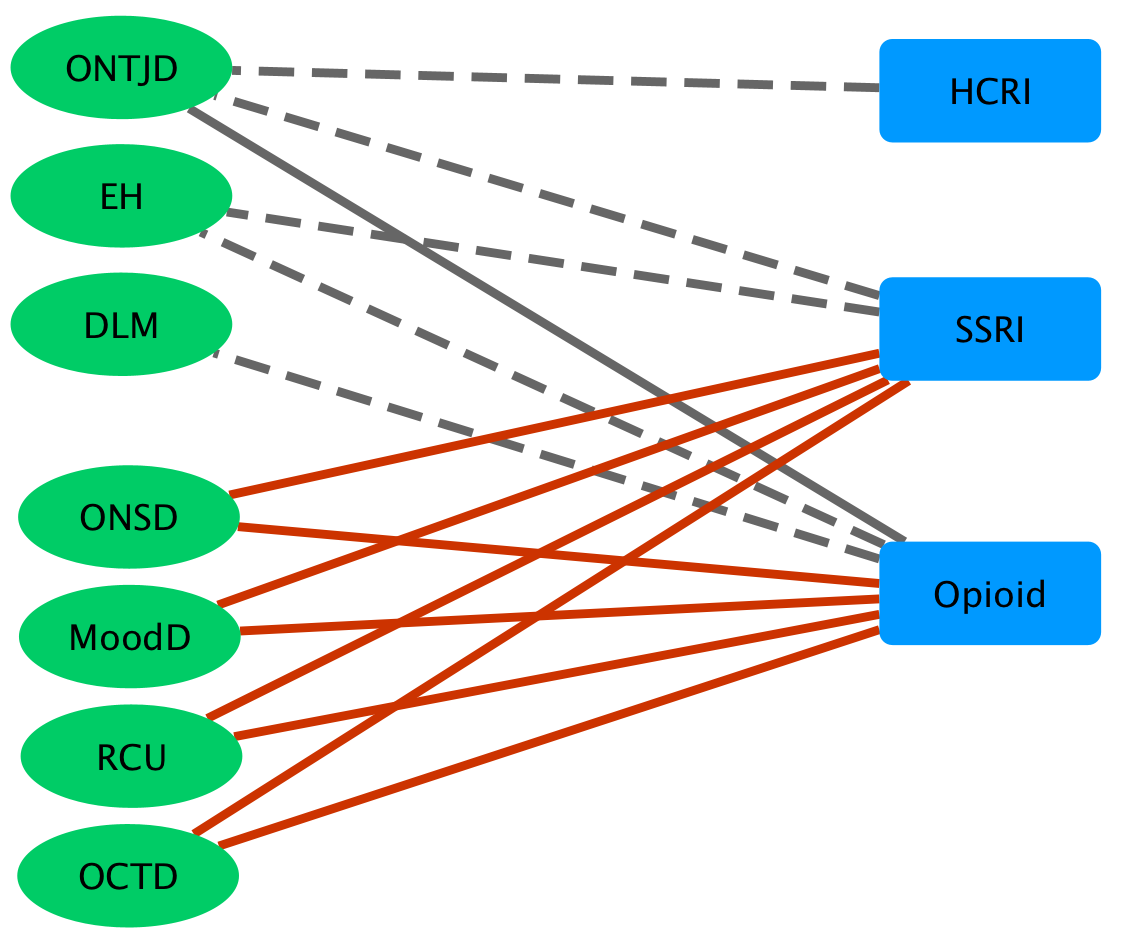}
		\caption{A pair of OUD-positive and OUD-negative samples that have different diagnoses but similar medication records.}
		\label{fig:attentions_md_1}
	\end{subfigure}\hfill%
	\begin{subfigure}{.33\linewidth}
		\centering
		\includegraphics[width=.7\linewidth]{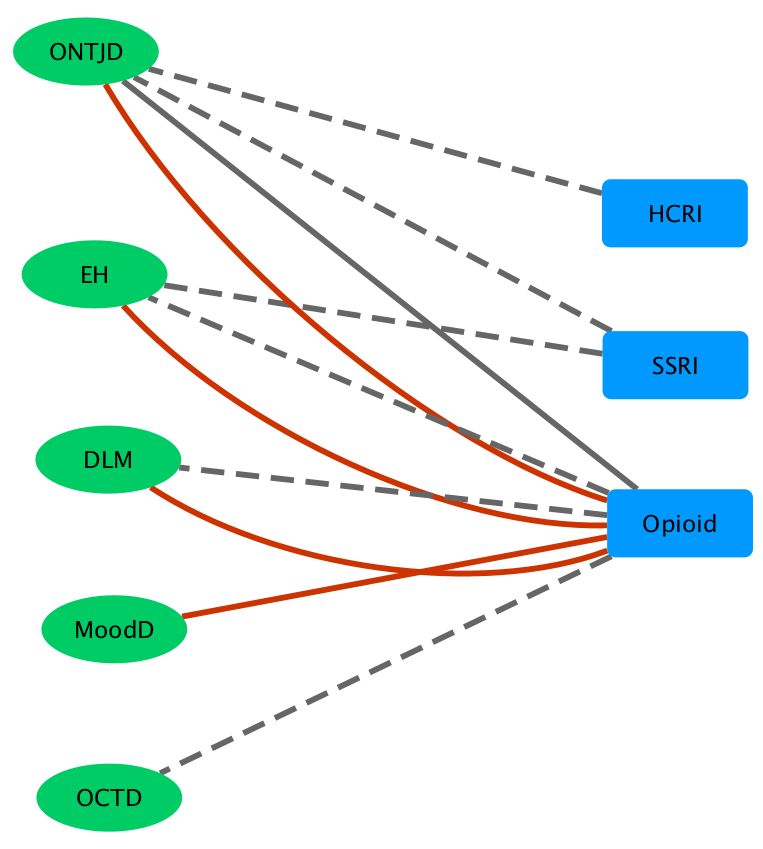}
		\caption{A pair of OUD-positive and OUD-negative samples that have very similar diagnoses and medication records.}
		\label{fig:attentions_md_2}
	\end{subfigure}
	\caption{Attention weights. Rectangular nodes represent medications and oval nodes represent diagnoses. Solid, dashed and dotted edges respectively mean strong, moderate and weak connections. We used abbreviations for medications and diagnoses, and provided the full names in (a).}
	\label{fig:attentions_md}
\end{figure}

Figure~\ref{fig:attentions_md_1} shows the attention weights computed with MUPOD on one OUD-positive and one OUD-negative patient. The cosine similarities of the medication and diagnosis streams of the two patients are 0.85 and 0.27, respectively, indicating that they have different diagnoses but similar medication records. The connections belonging to the positive and negative patients are well separated. Besides, almost all the strong connections are from the OUD-positive patient, while all the moderate connections are from the OUD-negative patient. Similarly, Figure~\ref{fig:attentions_md_2} shows the attention weights on one OUD-positive and one OUD-negative patient. The cosine similarities of the medication and diagnosis streams of the two patients are 0.71 and 0.93, respectively, indicating that they have very similar diagnoses and medication records. Although they have similar records and similar connections between medication and diagnoses nodes, the strengths of attention are different for the OUD-positive patient versus the OUD-negative patient and MUPOD was able to correctly classify these two samples. Note that ONTJD and Opioid are collected with both the OUD-positive link (red) and the OUD-negative link (black), indicating the ONTJD-Opioid  is often observed on both cases. Figure~\ref{fig:attentions_md} shows that the attention weights in MUPOD can be used to: 1) discriminate OUD-positive from OUD-negative patients and 2) reveal the relationships between medications that the patient has been prescribed with and the diagnoses he/she has been diagnosed with. These attention weights can further be accumulated across all patients in the cohort to create more generalized conclusions and OUD risk factor identification.


\section*{Conclusion}
OUD  is  a  public  health  crisis  costing  the  US  billions of dollars  annually in  healthcare, lost workplace productivity, and crime. In this study, we developed a multi-stream transformer model to analyze the long-term impact of medication application pattern, diagnosis history and demographic information, and to explore the associations within and between these streams of patients\textquotesingle \,data.  Our proposed model was able to predict the onset of OUD more effectively compared to baseline models including RF, SVM, LSTM and original transformer model. We discovered that the associations between medication and diagnosis streams are key factors that improve power to predict the development of subsequent OUD.

There are some limitations in our approach. First, the current model relies on patient demographic information and limited subset of medications and diagnoses as features. Incorporating more detailed diagnostic and medication information such as daily dose of opioid could refine the relationship between medications and diagnoses, and create more accurate OUD identification tools. Furthermore, this work only considered a cohort of 196,246 OUD patients who has been diagnosed with the OUD ICD9 or ICD10 codes at least once, ignoring all the undiagnosed OUD patients. For example, more than 224K patients in Truven have been prescribed with Buprenorphine or Methadone but without having any OUD diagnoses. These patients may be undiagnosed OUD patients and could be included in our future work. Second, the current approach cannot predict/estimate risks because the medication application patterns and the diagnosis history of patients that may lead to the increment of OUD risk has not been studied. Third, the explainability of MUPOD was explored using a few representative samples. However, more analysis and correlation analysis using more sophisticated methods such as heatmaps are needed in the future to interpret the model more efficiently. In the future, we will extend our model to address the aforementioned problems such as incorporating more medication and diagnosis features as well as the Morphine Milligram Equivalent (MME) information in MUPOD. The rationale is, given a patient who is constantly on the same type of medication for a while, the variation of the dosage may indicate whether the medication is still effective for the patient.

Despite the limitations of the model, the current approach adds detail to our understanding of the factors that may be important to the development of OUD. Our hope is that a more thorough understanding of the relationships between medications and diagnosis will eventually enable clinicians to identify individuals at risk for OUD at an earlier stage, and ideally, perhaps even prevent OUD.


\section*{Acknowledgments}
This research is supported by Kentucky Lung Cancer Research (grant no.KLCR-3048113817).

\small
\bibliographystyle{unsrt}
\setlength\itemsep{-0.1em}
\bibliography{thesis.bib} 


\end{document}